\documentclass{article}
\usepackage{ijcai17}
\usepackage{algorithm}
\usepackage{algorithmic}
\usepackage{amsmath}
\usepackage{cases}
\usepackage{times}
\usepackage{graphicx}

\newtheorem{theorem}{Theorem}[section]
\newtheorem{lemma}[theorem]{Lemma}

\usepackage{times}

\title{Solving a New 3D Bin Packing Problem with Deep Reinforcement Learning Method}
\author{Haoyuan Hu, Xiaodong Zhang, Xiaowei Yan, Longfei Wang, Yinghui Xu\\ 
Artificial Intelligence Department, Zhejiang Cainiao Supply Chain Management Co., Ltd., Hangzhou, China  \\
haoyuan.huhy@cainiao.com, yuhui.zxd@cainiao.com, xiaowei.yanxw@cainiao.com, \\
shouchu.wlf@cainiao.com, renji.xyh@taobao.com}

\begin{document}

\maketitle

\begin{abstract}
In this paper, a new type of 3D bin packing problem (BPP) is proposed, in which a number of cuboid-shaped items must be put into a bin one by one orthogonally. The objective is to find a way to place these items that can minimize the surface area of the bin. This problem is based on the fact that there is no fixed-sized bin in many real business scenarios and the cost of a bin is proportional to its surface area. Our research shows that this problem is NP-hard. Based on previous research on 3D BPP, the surface area is determined by the sequence, spatial locations and orientations of items. Among these factors, the sequence of items plays a key role in minimizing the surface area. Inspired by recent achievements of deep reinforcement learning (DRL) techniques, especially Pointer Network, on combinatorial optimization problems such as TSP, a DRL-based method is applied to optimize the sequence of items to be packed into the bin. Numerical results show that the method proposed in this paper achieve about 5\% improvement than heuristic method.  
\end{abstract}

\section{Introduction}

Bin packing problem (BPP) is a classical and important optimization problem in logistic system and production system. There are many variants of BPP, but the most meaningful and challenging one is 3D BPP, in which a number of cuboid-shaped items with different sizes should be packed into bins orthogonally. The size and cost of bins are fixed and known and the objective is to minimize the number of bins used, i.e., minimize the total cost. BPP is a typical and interesting combinatorial optimization problem and is NP-hard (~\cite{coffman1980performance}), so it is a very popular research direction in optimization area. In addition, BPPs have many applications in practice. An effective bin packing algorithm means the reduction of computation time, total packing cost and increase in utilization of resources.

Because the cost of packing materials, which is mainly determined by their surface area,  occupies the most part of packing cost, and we have found that in many real business scenarios there is no bin with fixed size (e.g., flexible and soft packing materials, not cartons or other bins, are used to pack items in cross-border e-commerce), so a new type of 3D BPP is proposed in our research. The objective of this new type of 3D BPP is to pack all items into a bin with minimized surface area. 

Due to the difficulty of obtaining optimal solutions of BPPs, many researchers have proposed various approximation or heuristic algorithms. To achieve good results, heuristic algorithms have to be designed specifically for different type of problems or situations, so heuristic algorithms have limitation in generality. In recent years, artificial intelligence, especially deep reinforcement learning, has received intense research and  achieved amazing results in many fields. In addition, DRL method has shown huge potential to solve combinatorial optimization problems (~\cite{vinyals2015pointer}, ~\cite{bello2016neural}). In this paper, a DRL-based method is applied to solve this new type of 3D BPP and numerical experiments based on real data are designed and conducted to demonstrate effectiveness of this method.

\section{Related Work}

\subsection{3D bin packing problem}

Bin packing problem is a classical and popular optimization problem. Since 1970s, it has attracted great interest of many researchers and some valuable achievements have been obtained. The two-dimensional BPP is NP-hard (~\cite{coffman1980performance}), so as a generalization of 2D BPP, 3D BPP is strongly NP-hard. For this reason, a lot of research focuses on approximation algorithms and heuristic algorithms. ~\cite{scheithauer1991three} proposed the first approximation algorithm for 3D BPP and investigated the performance bound of the algorithm. And many effective heuristic algorithms, such as Tabu Search (~\cite{lodi2002heuristic}, ~\cite{crainic2009ts}), guided local search (~\cite{faroe2003guided}), extreme point-based heuristics (~\cite{crainic2008extreme}), hybrid genetic algorithm (~\cite{kang2012hybrid}), have been proposed. There are also some research about exact solution method for 3D BPP. ~\cite{chen1995analytical} considered a problem of loading containers with cartons of non-uniform size, which is a generalization of 3D BPP where bins may have different sizes, and a mixed integer programming model was developed to obtain optimal solutions. An exact branch-and-bound algorithm for 3D BPP was proposed in ~\cite{martello2000three} and many instances with up to 90 items can be solved to optimality within a reasonable time limit. Some variants of BPP from real world are also studied, such as variable size bin packing problem (~\cite{kang2003algorithms}), bin packing problem with conflicts (~\cite{khanafer2010new}, ~\cite{gendreau2004heuristics}) and bin packing problem with fragile objects (~\cite{clautiaux2014lower}). 

Another class of packing problem, named strip packing problem, is also worth mentioning here, because it is very similar to our proposed problem. In the strip packing problem, a given set of cubiod-shaped items should be packed into a given strip orthogonally by minimizing the height of packing. The length and width of the strip is fixed and limited, and the height is infinite (for two dimensional strip packing problem, the width of strip is fixed and the length is infinite). This type of problem has many applications in steel industry and textile industry, and different types of algorithms have been proposed to solve the problem, such as exact algorithms in ~\cite{martello2003exact} and ~\cite{kenmochi2009exact}, approximation algorithm in ~\cite{steinberg1997strip}, heuristic algorithm in ~\cite{bortfeldt2007heuristic} and meta-heuristic algorithms in ~\cite{bortfeldt2006genetic} and ~\cite{hopper2001review}.

\subsection{DRL in combinatorial optimization}

Even though machine learning and combinatorial optimization have been studied for decades respectively, there are few investigations about application of machine learning method in combinatorial optimization problems. One research direction is designing hyper-heuristics based on reinforcement learning (RL) ideas. An overview of hyper-heuristics is presented in \cite{burke2013hyper}, in which some hyper-heuristics based on learning mechanism are discussed. In ~\cite{nareyek2003choosing}, the heuristics selection probability is updated based on non-stationary RL. In addition, various score updating methods have been proposed in the area of hyper-heuristics, such as binary exponential backoff (\cite{remde2009binary}), tabu search (\cite{burke2003tabu}) and choice function (\cite{cowling2000hyperheuristic}).


Recent advances in sequence-to-sequence model (\cite{sutskever2014sequence}) have motivated the research about neural combinatorial optimization. Attention mechanism, which is used to augment neural networks, contributes a lot in areas such as machine translation (\cite{bahdanau2014neural}) and algorithm-learning (\cite{graves2014neural}). In ~\cite{vinyals2015pointer}, a neural network with a specific attention mechanism named Pointer Net was proposed and a supervised learning method is applied to solve the Traveling Salesman Problem. ~\cite{bello2016neural} developed a neural combinatorial optimization framework with RL, and some classical problems, such as Traveling Salesman Problem and Knapsack Problem are solved in this framework. Because of the effectiveness and generality of the methodology proposed in ~\cite{bello2016neural}, our research is mainly based on their framework and methods.

\section{Deep Reinforcement Learning Method for 3D Bin Packing Problem}

\subsection{Definition of the problem}

In a typical 3D BPP, a set of items must be packed into fixed-sized bins in the way that minimizes the number of bins used. Unlike typical BPP with fixed-sized bins, we focus on the problem of designing the bin with least surface area that could pack all the items. In real business scenarios, such as cross-board e-commerce, no fixed-sized bin is available and flexible and soft materials are used to pack all the items. At the same time, the cost of a bin is directly proportional to its surface area. In this case, minimizing the surface area for the bin would bring great economic benefits. 

The exact formulation of our problem is given below. Given a set of cuboid-shaped items and each item $i$ is characterized by length($l_i$), width($w_i$) and height($h_i$). Our target is to find out the least surface area bin that could pack all items. We define $(x_i,y_i,z_i)$ as the left-bottom-back (LBB) coordinate of item $i$ and define $(0,0,0)$ as the left-bottom-back coordinate of the bin. The details of decision variables are shown in Table 1. Based on the descriptions of problem and notations, the mathematical formulation for the new type of 3D BPP is presented as follows:

\begin{equation*}
\begin{split}
&\min\,\,  L \cdot W + L \cdot H + W \cdot H\\ 
&\texttt{s}.\texttt{t}.\quad
\begin{cases}
s_{ij}+ u_{ij} + b_{ij}  =1  								   & (1)	\\ 
{\delta}_{i1} + {\delta}_{i2} + {\delta}_{i3} + {\delta}_{i4} + {\delta}_{i5} + {\delta}_{i6} =1  & (2) 	\\ 
x_i - x_j + L  \cdot  s_{ij}  \le  L - \hat{l_i}   										    & (3)	\\ 
y_i - y_j + W \cdot u_{ij} \le W - \hat{w_i} 										   & (4)	 \\ 
z_i - z_j + H \cdot b_{ij} \le  H - \hat{h_i}                                                                                   & (5)    \\ 
0 \le x_i \le L- \hat{l_i}             												& (6)\\ 
0 \le y_i \le W - \hat{w_i}         												& (7)\\ 
0 \le z_i \le H - \hat{h_i}           												& (8)\\ 
\hat{l_i} = {\delta}_{i1}   l_i + {\delta}_{i2}  l_i + {\delta}_{i3}  w_i + {\delta}_{i4}  w_i + {\delta}_{i5}  h_i + {\delta}_{i6}  h_i   & (9)\\   
\hat{w_i} = {\delta}_{i1}  w_i +{\delta}_{i2}  h_i +{\delta}_{i3}  l_i +{\delta}_{i4}  h_i +{\delta}_{i5}  l_i +{\delta}_{i6}  w_i      & (10)\\ 
\hat{h_i} =  {\delta}_{i1}  h_i +{\delta}_{i2}  w_i +{\delta}_{i3}  h_i +{\delta}_{i4}  l_i + {\delta}_{i5}  w_i +{\delta}_{i6}  l_i     & (11)\\
s_{ij} , u_{ij},  b_{ij}  \in \{0,1\}                                                                                                                      & (12)\\
 {\delta}_{i1}, {\delta}_{i2}, {\delta}_{i3}, {\delta}_{i4}, {\delta}_{i5}, {\delta}_{i6} \in \{0,1\}                                                     &(13)\\
\end{cases}
\end{split}
\end{equation*}

where $s_{ij} = 1$ if item $i$ is in the left side of item $j$, $u_{ij} = 1$ if item $i$ is under item $j$, $b_{ij} = 1$  if item $i$  is in the back of item $j$, ${\delta}_{i1} = 1$ if the orientation of item $i$ is front-up, ${\delta}_{i2}=1$ if the orientation of item $i$ is front-down, ${\delta}_{i3}=1$ if the orientation of item $i$ is side-up, ${\delta}_{i4}=1$ if the orientation of item $i$ is side-down, ${\delta}_{i5}=1$ if orientation of item $i$ is buttom-up, ${\delta}_{i6}=1$ if orientation of item $i$ is buttom-down.

Constraints $(9),(10),(11)$ denote the length, width, height of item $i$ after orientating it. Constraints $(1),(3),(4),(5)$ are used to guarantee there is no overlap between two packed items while constraints $(6),(7),(8)$ are used to guarantee the item will not be put outside the bin. 

We have tried to solve the problem by optimization solvers, such as IBM Cplex Optimizer, but it is very difficult to solve in reasonable time limit and we will prove this problem is NP-hard in the appendix.

\begin{table}[!htbp]
\scriptsize
\centering
\caption{Decision Variables}
\label{variable table}
\begin{tabular}{|l|l|l|}
\hline
Variable           &  Type              & Meaning \\ \hline
$L$			& Continuous   & the length of the bin                                                                 \\ \hline   
$W$			& Continuous   & the width of the bin								 \\ \hline  
$H$			& Continuous   & the height of the bin 							\\ \hline
$x_{i}$             &   Continuous  & LBB coordinate of item $i$ in $x$ axis        \\ \hline         
$y_i$               &   Continuous  &  LBB coordinate of item $i$ in $y$ axis                                                 \\ \hline
$z_i$               &   Continuous   & LBB coordinate of item $i$ in $z$ axis             \\ \hline
$s_{ij}$	      &    Binary	&  item $i$ is in the left side of item $j$ or not                        \\ \hline
$u_{ij}$	      &   Binary         &  item $i$ is under item $j$ or not                                    \\ \hline
$b_{ij}$	      &    Binary 	&  item $i$  is in the back of item $j$	or not		\\ \hline
${\delta}_{i1}$ &   Binary		&  orientation of item $i$ is front-up or not              \\ \hline
${\delta}_{i2}$ &   Binary		&  orientation of item $i$ is front-down or not                       \\ \hline
${\delta}_{i3}$ & Binary		&  orientation of item $i$ is side-up or not                      \\ \hline
${\delta}_{i4}$ & Binary		&  orientation of item $i$ is side-down or not                   \\ \hline
${\delta}_{i5}$ & Binary		&  orientation of item $i$ is buttom-up  or not                      \\ \hline
${\delta}_{i6}$ & Binary		&  orientation of item $i$ is buttom-down or not                       \\ \hline
\end{tabular} 
\end{table}

\subsection{A DRL-based method}

In this section, we will describe the DRL-based method to solve this new type of 3D BPP. Since solving it exactly is intractable, we use a constructive approach, i.e., packing items one by one in sequence. There are three class of decisions to make:

\begin{enumerate}
	\item the sequence in which the items are packed into the bins.	
	\item item orientation to be put into the bin.
	\item the strategy that selects an empty maximal space to put the item.
\end{enumerate}

We design a heuristic algorithm to choose the sequence, orientation and empty maximal space. When putting an item, the algorithm will go over all empty maximal spaces and 6 orientations for this item and choose the empty maximal space and orientation that yields least surface area. After that, we will go over all the remaining items and identify one that will yield least waste space. The detailed algorithm is described in the appendix. In this paper, DRL is used to find better sequence to pack the items, other strategies for choosing item orientation and empty maximal space are the same as the heuristic mentioned above. In doing so, we are only demonstrating that DRL can be powerful in finding a better solution than well-designed heuristic. In our future work, we will investigate how to incorporate all of item sequence, orientation and empty maximal space choice into DRL framework.	

\subsubsection{Architecture of the network}

In our research, the design of network architecture is inspired by the work of ~\cite{vinyals2015pointer} and ~\cite{bello2016neural}. In their studies, a neural network architecture named Pointer Net (Ptr-Net) is proposed and used to solve some classical combinatorial optimization problems, such as Traveling Salesman Problem (TSP) and Knapsack Problem. For example, when solving TSP, the coordinates of points on two-dimensional plane are used as input to the model step by step, and the sequence in which points are visited is the predicted results. This architecture is similar to sequence-to-sequence model, which is proposed in ~\cite{sutskever2014sequence} and is a powerful method in machine translation. There are two main differences between Ptr-Net and sequence-to-sequence model: first, the number of target classes in each step of the output in sequence-to-sequence model is fixed, but in Ptr-Net, the output dictionaries size is variable; second, the attention mechanism is used to blend hidden units of the encoder to a context vector in sequence-to-sequence model, but Ptr-Net use attention as a pointer to select a member of the input sequence as the output.

The neural network architecture in our research is shown in Figure 1. The input to this network is a sequence of size data (length, width and height) of items to be packed, and the output of this network is another sequence which represents the order we pack those items. The network consists two RNNs: an encoder network and a decoder network. At each step of encoder network, the size data (length, width and height) of one item are embedded and given as input to the LSTM cell and the cell output is collected. After the final step of the encoder network, the cell state and outputs are given to the decoder network. At each step of decoder network, one of the outputs of encoder network is selected as the input of the next step. For example, as show in Figure 1, the output of the 3rd step of decoder network is 4, so the output of the 4th step of encoder network is selected (pointed) and given as the input to the 4th step of the decoder network. And the attention mechanism and glimpse mechanism proposed in ~\cite{bello2016neural} is also used to integrate the information of output of decoder cell and outputs of encoder network to predict which item will be selected in each step.

\begin{figure}[!hbt]
\centering
\includegraphics[height=130pt]{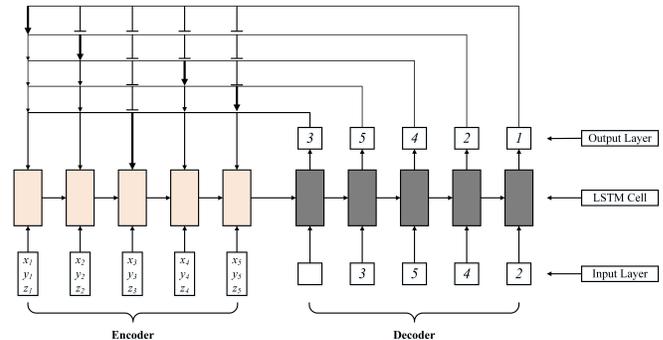}
\caption{Architecture of the neural network}
\end{figure}

\subsubsection{Policy-based reinforcement learning method}

In this paper, reinforcement learning methodology is used to train the neural network. The input of network can be denoted as $s = \{(l_i, w_i, h_i)\}_{i=1}^n$, where $l_i, w_i, h_i$ represents the length, width and height of the $i$th item respectively. The output of network is the sequence in which the items are packed into the bin, which can be denoted as $o$. And if the items are packed in this sequence, there will be a smallest bin that can pack all the items. We use the surface area (SA) of the bin to evaluate the sequence, and we use $SA(o|s)$ to denote the surface area. The stochastic policy of the neural network can be defined as $p(o|s)$, i.e., the probability of choosing sequence $o$ in which items are packed given a number of items $s$. And the goal of training is to give high probabilities to sequences that correspond to small surface areas. To be more specific, we use $\pmb{\theta}$ to denote the parameters of the neural network, and the training objective is the expected surface area, which is defined as:

\begin{equation*}
J(\pmb{\theta} |s) = E_{o \sim p_{\pmb{\theta}(\cdot | s)}}  SA(o|s)
\end{equation*}

In ~\cite{williams1992simple}, a general class of associative reinforcement learning algorithms, called REINFORCE algorithms are proposed. These algorithms can make weight adujustments in a direction that lies along the gradient of expected reinforcement. Based on the ideas of these algorithms, in each step of training, if the reward, baseline value and probability distribution of prediction are obtained, then the parameters of the neural network, $\pmb{\theta}$,  is incremented by an amount 

\begin{equation*}
\nabla_{\pmb{\theta}} J(\pmb{\theta}| s) = E_{o \sim p_{\pmb{\theta}(\cdot | s)}} [(SA(o | s) - b(s)) \nabla_{\theta} log p_{\pmb{\theta}} (o | s)]
\end{equation*}

where $b(s)$ denotes the baseline value of surface area and is used to reduce the variance of the gradients. And if we randomly get $M$ $i.i.d.$ samples $s_1, s_2, \dots, s_M$, then the above gradients can be approximated by:

\begin{equation*}
\nabla_{\pmb{\theta}} J(\pmb{\theta}| s) \approx \frac{1}{M} \sum_{i=1}^{M} [(SA(o_i | s_i) - b(s_i)) \nabla_{\theta} log p_{\pmb{\theta}} (o_i | s_i)]
\end{equation*}

\subsubsection{Baseline iteration: memory replay}

For a sample $s_i$, the baseline value $b(s_i)$ is initialized by calculating the surface area of a packing plan which is generated by a heuristic algorithm. And in each step, the baseline value is updated as:

\begin{equation*}
b'(s_i) = b(s_i) + \alpha (SA(o_i | s_i) - b(s_i))
\end{equation*}

where $SA(o_i | s_i)$ is the surface area calculated at each step. 

\subsubsection{Random sampling and beam search}

For each sample, the output of neural network is randomly sampled based on the probability distribution given by the policy network during training stage. While in testing stage, the greedy strategy is applied, i.e., in each step, the prediction with maximal probability is selected as output. In addition, a beam search method is used in the testing procedure to enhance the performance of neural network, i.e., the predictions with top-k highest probability are selected and maintained in each step.

As a conclusion of the discussion above, the training procedure of the neural network can be shown in Algorithm 1.

\begin{algorithm}                  
	\caption{Training Procedure}       
	\label{alg-summary-drl} 
	\begin{algorithmic}[1]
		\STATE Training set $S$, number of training steps $T$, batch size $B$.
		\STATE Initialize Pointer Net params $\theta$.
		\STATE Initialize baseline value according to heuristic algorithm.
		\FOR {\emph{t = 1 to T}}
			\STATE Select a batch of sample $s_i$ for $i \in \{ 1, \cdots, B   \}$.
			\STATE Sample solution $o_i $ based on $p_{\theta}(\cdot | s_i)$ for $i \in \{ 1, \cdots, B   \}$. 
			\STATE Let $g_{\theta} $= $\frac{1}{B} \sum_{i=1}^{B} [(SA(o_i | s_i) - b(s_i)) \nabla_{\theta} log p_{\pmb{\theta}} (o_i | s_i)]$.
			\STATE Update $\theta = ADAM(\theta, g_{\theta})$.
			\STATE Update baseline $b(s_i) = b(s_i) + \alpha (SA(o_i | s_i) - b(s_i))$ for $i \in \{ 1, \cdots, B   \}$. 
		\ENDFOR
		\STATE return pointer net parameters $\theta$.
	\end{algorithmic}
\end{algorithm}

\section{Experiments}
To test the performance of the model, a series of experiments on real data are conducted. The experiments can be classified into three categories based on the number of items in one customer order, i.e., 8, 10 and 12. In all of the experiments, we use 150,000 train samples and 150,000 test samples. Despite the difference in item number, we use the same hyper-parameters to train the model. We use mini-batch of size 128 and LSTM cell with 128 hidden units. We train the model with Adam optimizer with initial learning rate of $10^{-3}$ and decay every 5000 steps by a factor of 0.96. All the parameters are initialized randomly in $[-0.08,0.08]$ and clip L2 norm of our gradients to 1.0. We use the surface area calculated by heuristic algorithm as initial baseline input and apply $\alpha =0.7$ during the baseline iteration. We use 1000,000 steps to train the model and it will take about 12 hours for Tesla M40 GPU machine. When testing, we use beam search (BS) of size 3. Model implementation with TensorFlow will be available soon. The performance indicator is average surface area (ASA).

The results of testing are shown in Table 2. Using beam search (BS), the proposed method achieves $4.89\%$, $4.88\%$, $5.33\%$ improvement than heuristic algorithm for Bin8, Bin10 and Bin12. Optimal sequences for $5000$ samples of Bin8 are obtained by exhaustive method, and the gap between results of heuristic algorithm and optimal solutions is about $10\%$, which means that RL BS results are very close to optimal sequences.
\begin{table}[]
\centering
\caption{Average surface area}
\label{my-label}
\begin{tabular}{|l|l|l|l|l|}
\hline
No. of bins          & Random  & Heuristic & RL Sampling & RL BS \\ \hline
8      &   44.70   & 43.97     & 41.82             & 41.82		   \\ \hline         
10    &   48.38    & 47.33     & 45.03             & 45.02                    \\ \hline
12    &   50.78   & 49.34     & 46.71             & 46.71                    \\ \hline
\end{tabular}
\end{table}

\section{Conclusion}

In this paper, a new type of 3D bin packing problem is proposed. Different from the classical 3D BPP, the objective of the new problem is to minimize the surface area of the smallest bin that can pack all items. Due to the complexity of the problem, it is very difficult to obtain optimal solution and heuristic algorithm may have the problem of lack of generality. Therefore, we apply the Pointer Net framework and and a DRL-based method to optimize the sequence of items to be packed. The model is trained and tested with a large number of real data. Numerical experiment results show that the DRL-based method outperforms a well-designed, effective heuristic algorithm significantly. Our main contributions include: firstly, a new type of 3D BPP is proposed; secondly, the DRL technique is firstly applied in solving bin packing problem. In the future research, we will focus on investigation of more effective network architecture and training algorithm. In addition, integrating the selection of orientation and empty maximal space into the architecture of neural network is also worthy of study. 

\section*{Acknowledgements}

We are extremely grateful to our colleagues, including Rong Jin, Shenghuo Zhu and Sen Yang from iDST of Alibaba, Qing Da, Shichen Liu, Yujing Hu from Search Business Unit of Alibaba, Lijun Zhu, Ying Zhang and Yujie Chen from AI Department of Cainiao, for their insight and expertise that greatly assisted the research and the presentation of the paper. We would also like to show our sincere appreciation to the support of Jun Yang and Siyu Wang from Alibaba Cloud in the implementation of DRL network in TensorFlow.

\bibliographystyle{named}
\bibliography{ijcai17}

\begin{thebibliography}{}

\bibitem[\protect\citeauthoryear{Bahdanau \bgroup \em et al.\egroup
  }{2014}]{bahdanau2014neural}
Dzmitry Bahdanau, Kyunghyun Cho, and Yoshua Bengio.
\newblock Neural machine translation by jointly learning to align and
  translate.
\newblock {\em arXiv preprint arXiv:1409.0473}, 2014.

\bibitem[\protect\citeauthoryear{Bello \bgroup \em et al.\egroup
  }{2016}]{bello2016neural}
Irwan Bello, Hieu Pham, Quoc~V Le, Mohammad Norouzi, and Samy Bengio.
\newblock Neural combinatorial optimization with reinforcement learning.
\newblock {\em arXiv preprint arXiv:1611.09940}, 2016.

\bibitem[\protect\citeauthoryear{Bortfeldt and
  Mack}{2007}]{bortfeldt2007heuristic}
Andreas Bortfeldt and Daniel Mack.
\newblock A heuristic for the three-dimensional strip packing problem.
\newblock {\em European Journal of Operational Research}, 183(3):1267--1279,
  2007.

\bibitem[\protect\citeauthoryear{Bortfeldt}{2006}]{bortfeldt2006genetic}
Andreas Bortfeldt.
\newblock A genetic algorithm for the two-dimensional strip packing problem
  with rectangular pieces.
\newblock {\em European Journal of Operational Research}, 172(3):814--837,
  2006.

\bibitem[\protect\citeauthoryear{Burke \bgroup \em et al.\egroup
  }{2003}]{burke2003tabu}
Edmund~K Burke, Graham Kendall, and Eric Soubeiga.
\newblock A tabu-search hyperheuristic for timetabling and rostering.
\newblock {\em Journal of heuristics}, 9(6):451--470, 2003.

\bibitem[\protect\citeauthoryear{Burke \bgroup \em et al.\egroup
  }{2013}]{burke2013hyper}
Edmund~K Burke, Michel Gendreau, Matthew Hyde, Graham Kendall, Gabriela Ochoa,
  Ender {\"O}zcan, and Rong Qu.
\newblock Hyper-heuristics: A survey of the state of the art.
\newblock {\em Journal of the Operational Research Society}, 64(12):1695--1724,
  2013.

\bibitem[\protect\citeauthoryear{Chen \bgroup \em et al.\egroup
  }{1995}]{chen1995analytical}
CS~Chen, Shen-Ming Lee, and QS~Shen.
\newblock An analytical model for the container loading problem.
\newblock {\em European Journal of Operational Research}, 80(1):68--76, 1995.

\bibitem[\protect\citeauthoryear{Clautiaux \bgroup \em et al.\egroup
  }{2014}]{clautiaux2014lower}
Fran{\c{c}}ois Clautiaux, Mauro Dell’Amico, Manuel Iori, and Ali Khanafer.
\newblock Lower and upper bounds for the bin packing problem with fragile
  objects.
\newblock {\em Discrete Applied Mathematics}, 163:73--86, 2014.

\bibitem[\protect\citeauthoryear{Coffman \bgroup \em et al.\egroup
  }{1980}]{coffman1980performance}
Edward~G Coffman, Jr, Michael~R Garey, David~S Johnson, and Robert~Endre
  Tarjan.
\newblock Performance bounds for level-oriented two-dimensional packing
  algorithms.
\newblock {\em SIAM Journal on Computing}, 9(4):808--826, 1980.

\bibitem[\protect\citeauthoryear{Cowling \bgroup \em et al.\egroup
  }{2000}]{cowling2000hyperheuristic}
Peter Cowling, Graham Kendall, and Eric Soubeiga.
\newblock A hyperheuristic approach to scheduling a sales summit.
\newblock In {\em International Conference on the Practice and Theory of
  Automated Timetabling}, pages 176--190. Springer, 2000.

\bibitem[\protect\citeauthoryear{Crainic \bgroup \em et al.\egroup
  }{2008}]{crainic2008extreme}
Teodor~Gabriel Crainic, Guido Perboli, and Roberto Tadei.
\newblock Extreme point-based heuristics for three-dimensional bin packing.
\newblock {\em Informs Journal on computing}, 20(3):368--384, 2008.

\bibitem[\protect\citeauthoryear{Crainic \bgroup \em et al.\egroup
  }{2009}]{crainic2009ts}
Teodor~Gabriel Crainic, Guido Perboli, and Roberto Tadei.
\newblock Ts 2 pack: A two-level tabu search for the three-dimensional bin
  packing problem.
\newblock {\em European Journal of Operational Research}, 195(3):744--760,
  2009.

\bibitem[\protect\citeauthoryear{Faroe \bgroup \em et al.\egroup
  }{2003}]{faroe2003guided}
Oluf Faroe, David Pisinger, and Martin Zachariasen.
\newblock Guided local search for the three-dimensional bin-packing problem.
\newblock {\em Informs journal on computing}, 15(3):267--283, 2003.

\bibitem[\protect\citeauthoryear{Gendreau \bgroup \em et al.\egroup
  }{2004}]{gendreau2004heuristics}
Michel Gendreau, Gilbert Laporte, and Fr{\'e}d{\'e}ric Semet.
\newblock Heuristics and lower bounds for the bin packing problem with
  conflicts.
\newblock {\em Computers \& Operations Research}, 31(3):347--358, 2004.

\bibitem[\protect\citeauthoryear{Graves \bgroup \em et al.\egroup
  }{2014}]{graves2014neural}
Alex Graves, Greg Wayne, and Ivo Danihelka.
\newblock Neural turing machines.
\newblock {\em arXiv preprint arXiv:1410.5401}, 2014.

\bibitem[\protect\citeauthoryear{Hopper and Turton}{2001}]{hopper2001review}
Eva Hopper and Brian~CH Turton.
\newblock A review of the application of meta-heuristic algorithms to 2d strip
  packing problems.
\newblock {\em Artificial Intelligence Review}, 16(4):257--300, 2001.

\bibitem[\protect\citeauthoryear{Kang and Park}{2003}]{kang2003algorithms}
Jangha Kang and Sungsoo Park.
\newblock Algorithms for the variable sized bin packing problem.
\newblock {\em European Journal of Operational Research}, 147(2):365--372,
  2003.

\bibitem[\protect\citeauthoryear{Kang \bgroup \em et al.\egroup
  }{2012}]{kang2012hybrid}
Kyungdaw Kang, Ilkyeong Moon, and Hongfeng Wang.
\newblock A hybrid genetic algorithm with a new packing strategy for the
  three-dimensional bin packing problem.
\newblock {\em Applied Mathematics and Computation}, 219(3):1287--1299, 2012.

\bibitem[\protect\citeauthoryear{Kenmochi \bgroup \em et al.\egroup
  }{2009}]{kenmochi2009exact}
Mitsutoshi Kenmochi, Takashi Imamichi, Koji Nonobe, Mutsunori Yagiura, and
  Hiroshi Nagamochi.
\newblock Exact algorithms for the two-dimensional strip packing problem with
  and without rotations.
\newblock {\em European Journal of Operational Research}, 198(1):73--83, 2009.

\bibitem[\protect\citeauthoryear{Khanafer \bgroup \em et al.\egroup
  }{2010}]{khanafer2010new}
Ali Khanafer, Fran{\c{c}}ois Clautiaux, and El-Ghazali Talbi.
\newblock New lower bounds for bin packing problems with conflicts.
\newblock {\em European journal of operational research}, 206(2):281--288,
  2010.

\bibitem[\protect\citeauthoryear{Lodi \bgroup \em et al.\egroup
  }{2002}]{lodi2002heuristic}
Andrea Lodi, Silvano Martello, and Daniele Vigo.
\newblock Heuristic algorithms for the three-dimensional bin packing problem.
\newblock {\em European Journal of Operational Research}, 141(2):410--420,
  2002.

\bibitem[\protect\citeauthoryear{Martello \bgroup \em et al.\egroup
  }{2000}]{martello2000three}
Silvano Martello, David Pisinger, and Daniele Vigo.
\newblock The three-dimensional bin packing problem.
\newblock {\em Operations Research}, 48(2):256--267, 2000.

\bibitem[\protect\citeauthoryear{Martello \bgroup \em et al.\egroup
  }{2003}]{martello2003exact}
Silvano Martello, Michele Monaci, and Daniele Vigo.
\newblock An exact approach to the strip-packing problem.
\newblock {\em INFORMS Journal on Computing}, 15(3):310--319, 2003.

\bibitem[\protect\citeauthoryear{Nareyek}{2003}]{nareyek2003choosing}
Alexander Nareyek.
\newblock Choosing search heuristics by non-stationary reinforcement learning.
\newblock In {\em Metaheuristics: Computer decision-making}, pages 523--544.
  Springer, 2003.

\bibitem[\protect\citeauthoryear{Remde \bgroup \em et al.\egroup
  }{2009}]{remde2009binary}
Stephen Remde, Keshav Dahal, Peter Cowling, and Nic Colledge.
\newblock Binary exponential back off for tabu tenure in hyperheuristics.
\newblock In {\em European Conference on Evolutionary Computation in
  Combinatorial Optimization}, pages 109--120. Springer, 2009.

\bibitem[\protect\citeauthoryear{Scheithauer}{1991}]{scheithauer1991three}
Guntram Scheithauer.
\newblock A three-dimensional bin packing algorithm.
\newblock {\em Elektronische Informationsverarbeitung und Kybernetik},
  27(5/6):263--271, 1991.

\bibitem[\protect\citeauthoryear{Steinberg}{1997}]{steinberg1997strip}
A~Steinberg.
\newblock A strip-packing algorithm with absolute performance bound 2.
\newblock {\em SIAM Journal on Computing}, 26(2):401--409, 1997.

\bibitem[\protect\citeauthoryear{Sutskever \bgroup \em et al.\egroup
  }{2014}]{sutskever2014sequence}
Ilya Sutskever, Oriol Vinyals, and Quoc~V Le.
\newblock Sequence to sequence learning with neural networks.
\newblock In {\em Advances in neural information processing systems}, pages
  3104--3112, 2014.

\bibitem[\protect\citeauthoryear{Vinyals \bgroup \em et al.\egroup
  }{2015}]{vinyals2015pointer}
Oriol Vinyals, Meire Fortunato, and Navdeep Jaitly.
\newblock Pointer networks.
\newblock In {\em Advances in Neural Information Processing Systems}, pages
  2692--2700, 2015.

\bibitem[\protect\citeauthoryear{Williams}{1992}]{williams1992simple}
Ronald~J Williams.
\newblock Simple statistical gradient-following algorithms for connectionist
  reinforcement learning.
\newblock {\em Machine learning}, 8(3-4):229--256, 1992.

\end{thebibliography}

\appendix

\section{3D Bin Packing Heuristic Algorithm}
The detailed \textbf{3D bin packing heuristic algorithm} is:

\begin{algorithm}                  
  \caption{3D Bin Packing Heuristic Algorithm}       
  \label{3D bin packing heuristic algorithm} 
  \begin{algorithmic}[1]
    \STATE Denote the set of $n$ items as I. Each item is of length $l_i$, height $h_i$ and width $w_i$.
    \STATE Initialize a sufficiently large bin(B) with length $L$, width $W$, height $H$ (We could set $L=W = H = \sum_{i=1}^n max(l_i,h_i,w_i)$).
    \STATE Initilize the set of remaining items $\hat{I} = I$ and the set of empty maximal spaces as $ES = \emptyset$.
    \FOR {\emph{t = 1 to n}}
      \IF{$t = 1$}
        \STATE  Select an item with largest surface area.
        \STATE  Put the item into the bin and generate 3 empty maximal spaces $ES1$. Update $ES = ES1$.
      \ELSE
        \STATE Select a item $i$ from set S according to least waste space heuristic(Algorithm \ref{Least Waste Space Heuristic}). Update $\hat{I}  \leftarrow  \hat{I} \setminus i$.
        \STATE Select an empty maximal space from $ES$ and decide the orientation according to Least surface Area Heuristic(Algorithm \ref{Least surface Area Heuristic}).
      \STATE Generate new empty maximal spaces($ES1$) and delete those that are intersected and overlapped($ES2$). Update $ES  \leftarrow ES  \bigcup ES1  \setminus ES2$.
      \ENDIF      
    \ENDFOR
    \STATE Return the surface area of the bin that could pack all items.
  \end{algorithmic}
\end{algorithm}

\begin{algorithm}                  
  \caption{Least Surface Area Heuristic}       
  \label{Least surface Area Heuristic} 
  \begin{algorithmic}[1]
    \STATE Denote the set of empty maximal space as $ES$, the set of orientations as $O$. 
    \STATE Initialize the least surface area for item $i$ as $LSA_i = 3 \cdot {(max(l_i,w_i,h_i) \cdot n)}^2$. 
    \STATE Initialize best empty maximal space $\hat{s}_{i} =null$, best orientation as $\hat{o}_i = null$.
      \FOR{each empty maximal space $s \in ES$}
        \FOR{each orientation $o \in O$}
          \STATE Calculate the surface area $SA_{i,s,o}$ after putting item $i$ in empty maximal space $s$ with orientation $o$.
            \IF{$SA_{i,s,o}  < LSA_i$}
              \STATE  Update $\hat{s}_{i} =s$, $\hat{o}_i = o$ and $LSA_i  \leftarrow SA_{i,s,o}  $.
            \ELSIF{$SA_{i,s,o}  =LSA_i$}
              \STATE  Apply the tie-breaking rule. (Selecting $s, o$ if and only if $min(length(s)-o(l_i),width(s)-o(w_i),height(s)-o(h_i))$ is less than $min(length(s_i)-o_i(l_i),width(s_i)-o_i(w_i),height(s_i)-o_i(h_i))$, where $length(s), width(s), height(s)$ represents the length, width, height of empty maximal space $s$ and $o(l_i), o(w_i), o(h_i)$ represents the length, width, height of item $i$ with orientation $o$.)
            \ENDIF
        \ENDFOR
      \ENDFOR
      \STATE Return $\hat{s}_{i}, \hat{o}_i$ for item $i$.
  \end{algorithmic}
\end{algorithm}


\begin{algorithm}                  
  \caption{Least Waste Space Heuristic}       
  \label{Least Waste Space Heuristic} 
  \begin{algorithmic}[1]
    \STATE Denote the set of remaining items as $\hat{I}$.
    \STATE Denote the volume of item $i$ as $V_i$.
    \STATE Initialize the best item $\hat{i} =null$. 
    \STATE Initialize the Least volume as $LV= {(max(l_i,w_i,h_i) \cdot n)}^3$.
    \FOR {\emph{each item $i \in \hat{I}$ }}
      \STATE Calculate the volume $V$ after packing item $i$ according to $\hat{s}_{i}, \hat{o}_i$(which is determined by Least surface Area Heuristic). Denote the least waste space of item $LWV_i = V-V_i$.
      \IF{$LWV_i   < LV$}
        \STATE  Update $LV = LWV_i $.
        \STATE  Update $\hat{i} = i$.
      \ENDIF
    \ENDFOR
    \STATE Return item $i$.
  \end{algorithmic}
\end{algorithm}

The heuristic algorithm uses both least surface area heuristic and least waste space heuristic while our DRL method only uses least surface area heuristic.

%
%
%
%

%
%
%
%
%
%
%

\section{NP-hardness of New Type of 3D BPP}
\begin{lemma} 
The new type of 3D BPP proposed in this paper is NP-hard.
\end{lemma}

\textbf{Proof}: 
First of all, we will prove the new type of 2D BPP is NP-hard. To show it is NP-hard, we will give a reduction of 1D Bin Packing Problem. 

Given a one-dimensional bin packing problem, it consists of $n$ items with integer size $w_1,\cdots,w_n$ and bins with integer capacity $W$. The objective is to minimize the number of bins used to pack all items.

To convert it into new type of 2D BPP, we assume that there are $n$ items with width $w_i$ and height $1/(n \cdot max({w_i}))$. And there is also a item with width $W$ and height  $W \cdot n  \cdot max(w_i)$, which is called as \emph{Base Item}. The new type of 2D BPP problem is to find the bin with least surface area to pack the generated $n+1$ items.

Without loss of generality, we assume the \emph{Base Item} is on the left-buttom of the bin.  Adding one item on the right side of the \emph{Base Item} yields the total surface area is increased by at least $ (W \cdot n \cdot max(x_i) )/( n \cdot max(w_i)) = W$. At the same time, even if all the items are added on the upper side of the \emph{Base Item}, the total increased area is at most  $W$. Thus, all the items will be put on the upper side of the \emph{Base Item}. 

Next, we will prove the length and width of the item will not be reversed. If reversing one item with width and length, the increased area is at least $W \cdot min(w_i)$ for this item. However, the increased area is at most $W$ for all items if no item is reversed. 

If we can find out a bin with least surface area to pack this $n+1$ items, we find out the least number of bins of capacity $W$ that can contain $n$ items of size $w_1, \cdots, w_n$. Therefore, if we can solve the new type of 2D BPP in polynomial time, the one-dimensional bin packing problem can be solved in polynomial time, which completes the proof that this new type of 2D BPP is NP-hard unless P = NP.

For the new type of 3D bin packing problem, we will add length $1/(n \cdot max({w_i}))^2$ for each item in the 2D case, which will ensure no item will be added on the length side. Proof is the same as the 2D case.


\end{document}